\relax
  \documentclass[letterpaper]{article}

  \usepackage{ijcai17}

  \usepackage{times}
  \usepackage{helvet}
  \usepackage{courier}
  \usepackage[utf8]{inputenc} % allow utf-8 input
  \usepackage[T1]{fontenc}    % use 8-bit T1 fonts, not compatible with aaai
  \usepackage{hyperref}       % hyperlinks
  \usepackage{url}            % simple URL typesetting
  \usepackage{booktabs}       % professional-quality tables
  \usepackage{amsfonts}       % blackboard math symbols
  \usepackage{nicefrac}       % compact symbols for 1/2, etc.
  \usepackage{microtype}      % microtypography
  \usepackage{verbatim}

  \usepackage{epsfig}
  \usepackage{graphicx}
  \DeclareGraphicsExtensions{.eps}
  
  \usepackage[small]{caption}
  \usepackage{subcaption}
  \usepackage{amssymb}
  \usepackage{amsmath}
  \usepackage{soul}
  \usepackage{xcolor}
  \usepackage[linesnumbered,ruled,vlined]{algorithm2e}
  \usepackage{amsthm}

  \def\*#1{\mathbf{#1}}
  \DeclareMathOperator*{\argminA}{arg\,min}
  
\begin{document}
  \title{Self-Paced Multitask Learning with Shared Knowledge}
  \author{
  Keerthiram Murugesan
  \and
  Jaime Carbonell\\ Carnegie Mellon University, Pittsburgh, PA, USA \\{\{kmuruges,jgc\}@cs.cmu.edu}
}
  \maketitle
  \begin{abstract}
  
  This paper introduces self-paced task selection to multitask learning, where instances from more closely related tasks are selected in a progression of easier-to-harder tasks, to emulate an effective human education strategy, but applied to multitask machine learning. We develop the mathematical foundation for the approach based on iterative selection of the most appropriate task, learning the task parameters, and updating the shared knowledge, optimizing a new bi-convex loss function.  This proposed method applies quite generally, including to multitask feature learning, multitask learning with alternating structure optimization, etc. Results show that in each of the above formulations self-paced (easier-to-harder) task selection outperforms the baseline version of these methods in all the experiments. 
  
  \end{abstract}
  
  %%%%%%%%%%%%%%%%%%%%%%%%%%%%%%%%%%%%%%%%%%%%%%%%%%%%%%%%%%%%%%%%%%%%%%%%%

  \section{Introduction}

  %%%%%%%%%%%%%%%%%%%%%%%%%%%%%%%%%%%%%%%%%%%%%%%%%%%%%%%%%%%%%%%%%%%%%%%%%

Self-paced learning, inspired by established human education principles, defines a new machine learning paradigm based on a curriculum defined dynamically by the learner ("self-paced") instead of a fixed curriculum set {\it a-priori} by a teacher. It is an iterative approach that alternatively learns the model parameters and selects easier instances at first, progressing to harder ones \cite{kumar2010self}. However, naive extension of self-paced learning to the multitask setting may result in intractable increases in the number of learning parameters and therefore inefficient use of shared knowledge among the tasks. Existing work in this area is not scalable and/or lacks sufficient generality to apply to  several multitask learning challenges \cite{li2016self}.

Not all tasks are equal. Some tasks are easy to learn and some tasks are complex, facilitated by previously learned tasks to solve it efficiently. For example, classification task of whether an image has a bird or not can be learned  by solving easier component tasks first such as \textit{Is there a wing?}, \textit{Is there a beak?}, \textit{Does it have feathers?}, etc. The knowledge learned from these previously learned easier tasks can be used to solve the complex tasks effectively and such shared knowledge plays an important role in transfer of information between these tasks. This phenomenon is more evident in many real-world data such as object detection, weather prediction, landmine detection, etc.

We introduce a new learning framework for multiple tasks that addresses the aforementioned issues. It starts with easier set of tasks, and gradually introduces more difficult ones to build the shared knowledge base. Our proposed method provides a natural way to specify the trade-off between choosing the easier tasks to update the shared knowledge and learning new tasks using the knowledge acquired from previously learned tasks.
Our proposed framework based on self-paced learning for multiple tasks addresses these three key challenges: 1) it embeds task selection into the model learning; 2) it gradually learns the shared knowledge at the system's own pace; 3) it is generalizable to a wider group of multitask problems.

We first briefly introduce the self-paced learning framework. Next, we describe our proposed approach for self-paced multitask learning with efficient learning of latent task weights. We give a probabilistic interpretation of these task weights, based on their training errors.
We apply our learning framework to a few popular multitask problems such as Multitask Feature Learning, Multitask Learning with Alternating Structure Optimization (ASO), Mean regularized Multitask Learning and show that self-paced multitask learning significantly improves the learning performance of the original problem. In addition, we evaluate our method against several algorithms for sequential learning of multiple tasks. 

  %%%%%%%%%%%%%%%%%%%%%%%%%%%%%%%%%%%%%%%%%%%%%%%%%%%%%%%%%%%%%%%%%%%%%%%%%
  
  \section{Background: Self-Paced Learning}
  
  %%%%%%%%%%%%%%%%%%%%%%%%%%%%%%%%%%%%%%%%%%%%%%%%%%%%%%%%%%%%%%%%%%%%%%%%%
  Given a set of $N$ training instances along with their labels $({x}_i,{y}_i)_{i \in [N]}$, the general form of the objective function for single task learning is given by:
  \begin{equation}
  \begin{aligned}
  \mathcal{E}_\lambda\{\hat{\mathbf{w}}\} = {\arg \min}_{\mathbf{w}} & \sum_{i \in [N]} \ell({y}_i,f({x}_i,\mathbf{w})) + \rho_\gamma(\*w) \\
  \end{aligned}
  \label{eq:stl}
  \end{equation}

  where $\rho_\gamma(\*w)$ is the regularization term on the model parameters and typically it is set to $\rho_\gamma(\*w) = \gamma ||\*w||_2^2$ (\textit{ridge} or L2 penalty) or $\gamma ||\*w||_1$ (\textit{lasso} or L1 penalty). $\gamma$ is the regularization parameter and $[N]$ is the index set $\{1,2, \ldots N\}$

  Self-paced learning (\textit{SPL})  provides a strategy for simultaneously selecting the easier instances and re-estimating the model parameters $\mathbf{w}$ at each iteration \cite{kumar2010self}. We assume a linear predictor function $f({x}_i,\mathbf{w})$ with unknown parameter $\mathbf{w}$.
  Self-paced learning solves the following objective function:

  \begin{equation}
  \begin{aligned}
  \mathcal{E}_\lambda\{\hat{\mathbf{w}},\hat{\mathbf{\tau}}\}  = \argminA_{\mathbf{w}, \mathbf{\tau} \in \varOmega}& \sum_{i \in [N]} \tau_{i} \ell({y}_i,f({x}_i,\mathbf{w}))\\
  & + \rho_\gamma(\*w) + \lambda r(\*\tau)\\
  \end{aligned}
  \label{eq:spl}
  \end{equation}
  where $r(\*\tau)$ is the regularization term, $\varOmega$ is the domain space of $\*\tau$,  $\rho_\gamma(\*w)$ is the regularization term on model parameters $\*w$ as defined earlier, and  $\lambda$ is the regularization parameter that identifies the difficulty of the instances. There are two unknowns in equation \ref{eq:spl}: model parameter vector $\mathbf{w}$ and the selection parameter $\mathbf{\tau}$ (restricted to the domain $\varOmega$).

  A common choice of the constraint space $\mathcal{C}=$ $\{\rho_\gamma(\*w), r(\*\tau),\varOmega\}$ in \textit{SPL} is   $\{ \gamma||\*w||_2^2, -||\*\tau||_1 \},\{0,1\}^N\}$. See \cite{jiang2015self} for more examples on the constraint space. With this setting, equation \ref{eq:spl} is a bi-convex optimization problem over $\mathbf{w}$ and $\mathbf{\tau}$, which can be efficiently solved by \textit{alternating minimization}. Given a fixed $\*\tau$, the solution for $\*w$ can be obtained using any off-the-shelf solver and for a fixed $\*w$, solution for $\*\tau$ can be given as follows:
  \begin{align} \label{eq:spltau}
      \hat{\tau_{i}} &=\begin{cases}
      1       & \quad \text{if }\ell({y}_i,f({x}_i,\mathbf{w})) < \lambda\\
      0  & \quad otherwise\\
    \end{cases} \quad \forall i \in [N]
  \end{align}
		  
  There exists an intuitive explanation for this alternative search strategy: 1) when updating $\*\tau$ with a fixed $w$, a sample whose loss is smaller than a certain threshold $\lambda$ is taken as an “easy” sample because it is a sample with ``\textit{less error}'', and will be selected in training ($\tau_i^*=1$) or otherwise unselected ($\tau_i^*=0$); 2) when updating $w$ with a fixed $\tau$, the classifier is trained only on the selected “easy” samples. When $\lambda$ is small, only “easy” samples with small losses will be considered.

  %%%%%%%%%%%%%%%%%%%%%%%%%%%%%%%%%%%%%%%%%%%%%%%%%%%%%%%%%%%%%%%%%%%%%%%%%

  \section{Self-Paced Multitask Learning with Shared Knowledge}

  %%%%%%%%%%%%%%%%%%%%%%%%%%%%%%%%%%%%%%%%%%%%%%%%%%%%%%%%%%%%%%%%%%%%%%%%%

Suppose we are given $T$ tasks where the $t$-th task is associated with $N_t$ training examples.
Denote by $\big\{ (x_i^{t}, y^{t}_i)\big\}_{i=1}^{N_t}$ 
and $\mathcal{L}(\*{y}_t,f(\*{X}_t,\mathbf{w}_t)) = \frac{1}{N_t} \sum_{i \in [N_t]} \ell({y}_{i}^t,f({x}_{i}^t,\mathbf{w}_t))$
the training set and loss for task $t$, respectively. In this paper, we consider a more general formulation for multitask learning, which is given by \cite{caruana1997multitask,baxter2000model,evgeniou2004regularized}:
  \begin{equation}
  \begin{aligned}
  \mathcal{E}_\lambda\{\hat{\mathbf{W}},\hat{\*\Theta}\}  = \argminA_{{\mathbf{W}, \*\Theta \in \*\Gamma}}& \sum_{t \in [T]}\mathcal{L}(\*y_t,f(\*X_t,\*w_t)) \\
  & + P_\gamma(\*W,\*\Theta)\\
  \end{aligned}
  \label{eq:mtl}
  \end{equation}
  
  where $P_\gamma(\*W,\*\Theta)$ is the regularization term on task parameters $\*W$, $\*\Theta$ is the knowledge shared among the tasks which depends on the problem under consideration.  We assume that $P_\gamma(\*W,\*\Theta)$ can be written as $\sum_{t \in [T]}P_\gamma(\*w_t,\*\Theta)$, such that, for a given $\*\Theta$, the above objective function decomposes into $T$ independent optimization problems.  $P_\gamma(\*w_t,\*\Theta)$ gives a scoring function on how easier the task is, compared to that of the learned knowledge $\*\Theta$. Several multitask learning problems fall under this general characterization. For example, Multitask Feature Learning (\textit{MTFL}), Regularized Multitask Learning (\textit{MMTL}), Multitask learning with manifold regularization (\textit{MTML}), Multitask learning via Alternating Structure Optimization (\textit{MTASO}), Sparse coding for multitask learning (\textit{SC-MTL}), etc \cite{evgeniou2007multi,evgeniou2004regularized,agarwal2010learning,ando2005framework,maurer2013sparse}. With this formulation, one can easily extend the $SPL$ framework to multitask setting, by considering instance weights for each task.

\begin{equation}
\begin{aligned}
\mathcal{E}_\lambda\{\hat{\mathbf{W}},\hat{\*\Theta},\hat{\mathbf{\tau}}\}  = \argminA_{\substack{\mathbf{W}, \*\Theta \in \*\Gamma\\ \mathbf{\tau} \in \varOmega}}& \sum_{t \in [T]}  \frac{1}{N_t} \sum_{i \in [N_t]} \tau_{ti}  \ell({y}_{i}^t,f({x}_{i}^t,\mathbf{w}_t)) \\
& + P_\gamma(\*W,\*\Theta) + \lambda r(\*\tau)\\
\end{aligned}
\label{eq:naive_spmtl}
\end{equation}
But there are two key issues with this naive extension of \textit{SPL}: 1) The above formulation fails to effectively utilize the knowledge shared among the tasks; 2) The number of unknown parameters $\*\tau$ grows with the total number of instances $N=\sum_t N_t$ from all the tasks. This is a serious problem especially when the number of tasks $T$ is large \cite{weinberger2009feature} and/or when manual annotation of task instances is expensive \cite{kshirsagar2013multitask}. 

To address these issues, we consider task-level weights, instead of instance-level weights. Our motivation behind this approach is based on the human educational process. When students learn a new concept, they (or their teachers) choose a new task that is relevant to their recently-acquired knowledge, rather that more distant tasks or concepts or other haphazard selections. Inspired by this interpretation, we propose the following objective function for Self-Paced Multitask Learning (\textit{sp}{MTL}): 

\begin{equation}
\begin{aligned}
\mathcal{E}_\lambda\{\hat{\mathbf{W}},\hat{\*\Theta},\hat{\mathbf{\tau}}\}  = \argminA_{\substack{\mathbf{W}, \*\Theta \in \*\Gamma\\ \mathbf{\tau} \in \varOmega}}& \sum_{t \in [T]} \tau_{t} \big[\mathcal{L}(\*y_t,f(\*X_t,\*w_t)) \\
& + P_\gamma(\*w_t,\*\Theta)\big] + \lambda r(\*\tau)\\
\end{aligned}
\label{eq:spmtl}
\end{equation}

Note that the number of parameters $\*\tau_t$ depends on $T$ instead of $N$ and the $\*\tau_t$ depends on both the training error of the task and the task regularization term for the shared knowledge $\*\Theta$. 

The pseudo-code is in Algorithm \ref{alg:spmtl}. The learning algorithm defines a task as "easy" task if it has low training error $\frac{1}{N_t} \sum_{i \in [N_t]} \ell({y}_{i},f({x}_{i},\mathbf{w}_t))$ and similar to the shared knowledge representation $P_\gamma(\*w_t,\*\Theta)$. These tasks will be selected in building the shared knowledge $\*\Theta$.  Following Equation \ref{eq:spltau}, we can define  $\tau_t$ as \footnote{For correctness of the algorithm, we set $\tau_t=\delta$ for the hard tasks, instead of $\tau_t=0$ with $\delta=0.01$.}:
\begin{align} \label{eq:spmtlsparse}
      \hat{\tau_{t}} &=\begin{cases}
      1       & \quad \text{if }\mathcal{L}(\*{y}_t,f(\*{X}_t,\mathbf{w}_t^{(k)})) \\
      & + P_\gamma(\*w_t^{(k)},\*\Theta^{(k-1)}) < \lambda\\
      \delta  & \quad otherwise\\
      \end{cases} \quad \forall t \in [T]
      \end{align}

  \begin{algorithm}[t]
      
      %\SetAlgoLined
      \SetKwInOut{Input}{Input}
      \SetKwInOut{Output}{Output}
      \Input{$\mathcal{D}=\{(\*X_t,\*y_t)\}_{t=1}^T, \*\Theta^{(0)}, \mathit{c}> 1$}
      \Output{$\*W, \*\Theta$}
      $k \gets 1, \lambda \gets \lambda_0$
      
      \Repeat{$\Vert \*\tau^{(k)} - \*\tau^{(k-1)} \Vert_2^2 \leq \epsilon$}
      {
	
	\textit{Solve} for $\*w^{(k)}_t \gets {\arg \min}_{\*w} \mathcal{L}(\*y_t,f(\*X_t,\*w)) + P_\gamma(\*w,\*\Theta^{(k-1)})$  $\forall t $ \;
	
	\textit{Solve} for $\*\tau^{(k)}$ using equation (\ref{eq:spmtlsparse}) or equation (\ref{eq:spmtlprob}) \;

       \textit{Solve} for $\*\Theta^{(k)}$ :
       
       $\*\Theta^{(k)} \gets {\arg \min}_{\*\Theta} \sum_{t \in [T]} \tau^{(k)}_t  P_\gamma(\*w_t^{(k)},\*\Theta)$\;
       
	  $\lambda \gets \mathit{c}\lambda$\;
	  $k \gets k + 1$\;
      }
      \caption{Self-Paced Multitask Learning: A General Framework}
      \label{alg:spmtl}
  \end{algorithm}

  \begin{comment}
	\begin{align} \label{eq:spmtlsparse}
      \hat{\tau_{t}}^{(k)} &=\begin{cases}
      1       & \quad \text{if }\mathcal{L}(\*{y}_t,f(\*{X}_t,\mathbf{w}_t^{(k)})) \\
      & + P_\gamma(\*w_t^{(k)},\*\Theta^{(k-1)}) < \lambda\\
      \delta  & \quad otherwise\\
      \end{cases} \quad \forall t \in [T]
      \end{align}
      
      $\mathcal{A} \gets supp(\*\tau^{(k)})$\;
      \textit{Solve} for $\*\Theta^{(k)} \gets {\arg \min}_{\*\Theta}  P_\gamma(\*W_\mathcal{A}^{(k)},\*\Theta)$\;
 \end{comment}
      
\begin{comment}
For example, setting $\mathcal{C}=\{ \gamma||\*w||_2^2,  -||\*\tau||_1 + \mu||\*\tau||_2^2 ,[0,1]^N\}$, the closed form solution for $\*\tau$ becomes:
\begin{equation}
\hat{\tau_{i}}= \frac{[\lambda - \ell({y}_i,f({x}_i,\mathbf{w}))     ]_+}{\sum_{i'}[\lambda - \ell({y}_{i'},f({x}_{i'},\mathbf{w}))      ]_+}, \quad \forall i \in [N]
\end{equation}
\end{comment}

For multitask setting, it is desirable to consider an alternative constraint space that gives probabilistic interpretation for $\*\tau$. By setting $\mathcal{C}=\{ \gamma||\*w||_2^2,  -\mathbf{H}\left( \*\tau \right) ,\Delta^{N-1}\}$ , we get
\begin{equation}
	\hat{\tau_{t}} \propto
	{\exp(-[\mathcal{L}(\*{y}_t,f(\*{X}_t,\mathbf{w}_t)) + P_\gamma(\*w_t,\*\Theta)]/\lambda)}, 
	\label{eq:spmtlprob}
\end{equation}

where $\mathbf{H}\left( \*\tau \right)=-\sum_{t \in [T]} \tau_t \log \tau_t$ denotes the entropy of the probability distribution $\*\tau$ over the tasks. The key idea is that the algorithm, at each iteration, maintains a probability distribution over the tasks to identify the simpler tasks based on the shared knowledge. Similar approach has been used in learning relationship between multiple tasks in an online setting \cite{murugesan2016adaptive}. Using this representation, we can use $\*\tau$ to sample, at each iteration, the "easy" tasks and thus makes the learning problem scalable using stochastic approximation when the number of tasks is large. 
It is worth noting that our framework can easily handle outlier tasks by a simple modification to Algorithm \ref{alg:spmtl}. Since outlier tasks are different from the main tasks and are usually difficult to learn, we can take advantage of this simple observation for early stopping, before the algorithm visits all the tasks \cite{romera2012exploiting}.

Our algorithm can be easily generalized to other types of updating rules by replacing $\exp$ in \eqref{eq:spmtlprob} with other functions. In latter cases, however, $\*\tau$ may no longer have probabilistic interpretations. Algorithm \ref{alg:spmtl} shows the basic steps in learning the task weights and the shared knowledge. The algorithm uses an additional parameter $'c'$ that controls the learning pace of the self-paced procedure. Typically, $'c'$ is set to some value greater than $1$ (in our experiments, we set it to $1.1$) such that, at each iteration, the threshold $\lambda$ is relaxed to included more tasks. The input to the algorithm also takes $\*\Theta^{(0)}$, initial knowledge about the domain and can be initialized based on some external sources.

\begin{comment}
  \subsection{Handling Outlier Tasks}
  
  Our framework can easily handle outlier tasks by a simple modification to Algorithm \ref{alg:spmtl}. Since outlier tasks are different from the main tasks and are usually difficult to learn, we can take advantage of this simple observation for early stopping, before the algorithm visits all the tasks (i.e., $\mathcal{A}=[T]$). Here, we introduce two early stopping criteria to achieve our goal. First, we consider the maximum value that the $\lambda$ can take: $\lambda_{\max}$. Second, we learn the joint $\*\Theta$ until our active set contains consider number of tasks in the pool: $|\mathcal{A}| < T$.
\end{comment}

  \subsection{Motivating Examples}
  We give three examples to motivate our self-paced learning procedure. We briefly discuss how our algorithm alters the learning pace of the original problem. Note that the existing implementation of these problems can be easily "self-paced", by simply adding a few lines of code to get a better performance of the original problem.  We refer the readers to \cite{evgeniou2007multi,agarwal2010learning,ando2005framework} for additional background. 
  
  \subsubsection{Example 1: Self-Paced Mean Regularized Multitask Learning (\textit{sp}MMTL)}
  Mean Regularized Multitask learning assumes that all task parameters are close to some fixed parameter $\*w_0$ in the parameter space. \textit{sp}MMTL learns $\*\tau$ to select the easy tasks based on the distance of each task parameter $\*w_t$ from $\*w_0$. 
  
  \begin{equation}
  \begin{aligned}
  \mathcal{E}_{\textit{MMTL},\lambda}  = \argminA_{\substack{\{\*w_1,\*w_2, \ldots \*w_T\}\\\mathbf{w}_0, \mathbf{\tau} \in \varOmega}}& \sum_{t \in [T]} \tau_{t} \mathcal{L}(\*y_t,f(\*X_t,\*w_t)) \\
  & + \gamma ||\*w_t-\*w_0||_2^2  + \lambda ||\*\tau||_1\\
  \end{aligned}
  \label{eq:spmmtl}
  \end{equation}
  
  In the above objective function, we can get the closed-form solution for $\*w_0$ as $\*w_0=\frac{1}{T}\sum_{t=1}^T \*w_t$ which is the mean of the task parameters.

  \textbf{Example 2: Self-paced Multitask Feature Learning (\textit{sp}MTFL)}
  Multitask feature learning  learns a common feature representation $\*D$ shared across multiple related tasks. In addition to learning the task parameters and the shared feature representation, \textit{sp}MTFL learns $\*\tau$ to select the easy tasks first, defined by the learning parameter $\lambda$. The algorithm starts with these easy tasks to learn the shared feature representation which is used for solving progressively harder tasks.
  \begin{equation}
  \begin{aligned}
  \mathcal{E}_{\textit{MTFL},\lambda}  = &\argminA_{\substack{\{\*w_1,\*w_2, \ldots \*w_T\}\\ \mathbf{D} \in \*S_{++}^d \\ \mathbf{\tau} \in \varOmega}} \sum_{t \in [T]} \tau_{t} \mathcal{L}(\*y_t,f(\*X_t,\*w_t)) \\
  & + \gamma \sum_{t \in [T]} \tau_t \langle \*w_t, \*D^{-1} \*w_t)   + \lambda r(\*\tau)\\
  \end{aligned}
  \label{eq:spmtfl}
  \end{equation}

  %where $\*D$ has a closed form solution: $\*D=\frac{(\*W\*W^\top)^{\frac{1}{2}}}{tr(\*W\*W^\top)^{\frac{1}{2}}}$. 
  
  The value of $\tau_t$  determines the importance of a task in learning this shared feature representation, i.e., tasks with high probability contributes more towards learning  $\*D$ than the tasks with low probability.
  
  \textbf{Example 3: Self-paced Multitask learning with Alternating Structure Optimization (\textit{sp}MTASO)}
  
Alternating Structure Optimization learns a shared low-dimensional predictive structure $\*U$ on a hypothesis space from multiple-related tasks. This low-dimensional structure along with the low-dimensional model parameters $\*v_t$ are learned gradually from easy tasks guided by $\*\tau$.
   \begin{equation}
  \begin{aligned}
  \mathcal{E}_{\textit{MTASO},\lambda}  = &\argminA_{\substack{\{\*w_1,\*w_2, \ldots \*w_T\} \\ \*U\*U^\top=I_{h\times h} \\ \mathbf{\tau} \in \varOmega}} \sum_{t \in [T]} \tau_{t} \mathcal{L}(\*y_t,f(\*X_t,\*w_t)) \\
  & + \gamma \sum_{t \in [T]} \tau_t ||\*w_t-\*U^\top \*v_t||_2^2  + \lambda r(\*\tau)\\
  \end{aligned}
  \label{eq:spmtaso}
  \end{equation}
  
    \begin{comment}
  \textbf{Example 3: Self-paced Multitask learning with Manifold regularization (\textit{sp}MTML)}
  Manifold-regularized multitask learning assumes that all the task parameters lie on a manifold $\mathcal{M}$. It alternatively learns the task parameters and the manifold. Our proposed learning algorithm \ref{alg:spmtl} chooses the tasks that are good representatives for constructing a reliable manifold structure. The objective function for \textit{sp}MTML is given as follows:
   \begin{equation}
  \begin{aligned}
  \mathcal{E}_{\textit{MTML},\lambda}  = \argminA_{\substack{\{\*w_1,\*w_2, \ldots \*w_T\} \\ \*W^{\mathcal{M}} \\ \mathbf{\tau} \in \varOmega}}& \sum_{t \in [T]} \tau_{t} \mathcal{L}(\*y_t,f(\*X_t,\*w_t)) \\
  & + \gamma \sum_{t \in [T]} \tau_t ||\*w_t-\*w^{\mathcal{M}}_t||_2^2  + \lambda r(\*\tau)\\
  \end{aligned}
  \label{eq:spmtml}
  \end{equation}

  where $\*w^{\mathcal{M}}_t=g(h(\*w_t))$ is the projection distance of $\*w_t$ from the manifold $\mathcal{M}$.
  \end{comment}
  
  %%%%%%%%%%%%%%%%%%%%%%%%%%%%%%%%%%%%%%%%%%%%%%%%%%%%%%%%%%%%%%%%%%%%%%%%%
  
  \section{Related Work}
  
  %%%%%%%%%%%%%%%%%%%%%%%%%%%%%%%%%%%%%%%%%%%%%%%%%%%%%%%%%%%%%%%%%%%%%%%%%
  In this section, we briefly review two learning methods that are most related to our proposed learning algorithm. Both these methods learn from multiple tasks sequentially in a specific order to either improve the learning performance or to speedup the algorithm.
  \citeauthor{pentina2015curriculum} (\citeyear{pentina2015curriculum}) propose a curriculum learning method (\textit{CL}) for multiple tasks to find the best order of tasks to be learned based on training error. The tasks are solved in a sequential manner based on this order by transferring information from the previously learned tasks to the next ones through shared task parameters. They show that this sequential learning of tasks in a meaningful order can be superior than solving the tasks simultaneously. The objective function of \textit{CL} for learning the best task order and the task parameters is given as follows:
   \begin{equation}
  \begin{aligned}
  \mathcal{E}_{\textit{CL}}  = \argminA_{\substack{\{\*w_1,\*w_2, \ldots \*w_T\} \\ \*\pi \in \Psi_T}}& \sum_{t \in [T]} \mathcal{L}(\*y_{\pi(t)},f(\*X_{\pi(t)},\*w_{\pi(t)})) \\
  & + \gamma \sum_{t \in [T]} ||\*w_{\pi(t)}-\*w_{\pi(t-1)}||_2^2  \\
  \end{aligned}
  \label{eq:clmtlobj}
  \end{equation}
  
  where $\Psi_T$ is the symmetric group of all permutations over $[T]$.
  Since, minimizing with respect to all possible permutations $\pi \in \Psi_T$ is an expensive combinatorial problem, they suggest a greedy, incremental procedure for approximating the task order. Their method shares with ours the motivation of learning from easier tasks first, and then gradually add more difficult tasks, based on training errors.  But unlike our proposed method, which utilizes shared knowledge from all previous tasks, their method does not allow sharing between different levels of task relatedness. In addition, the Euclidean distance based regularization in their objective function forces the parameter of newly learned task to be similar to its immediate predecessor. This more myopic approach can be a restrictive assumption for many applications.

  Perhaps the most relevant work to ours in the context of lifelong learning is from \cite{ruvolo2013ella}, which learns the shared basis $\*L$ from tasks that arrives sequentially. They propose an efficient online multitask learning algorithm (\textit{ELLA}) that allows the transfer of knowledge from previously learned tasks to the new tasks using this shared basis. The task parameters are represented as a sparse linear combination of the columns of the shared basis $\*w_t=\*L\*s_t$.  The motivation for \textit{ELLA} and our method are significantly different. Whereas \textit{ELLA} tries to achieve nearly identical to the performance of batch MTL with increased speedup in learning, our proposed method focuses on improving the learning performance over that of the original algorithm, with minimal changes to said original algorithm. Unlike our proposed method, \textit{ELLA} cannot be easily generalized to existing multitask problems. It only uses efficient update equations specific to their proposed objective function. 
  
   \begin{comment}
  \textit{ELLA} optimizes the following objective function:
   
  \begin{equation}
  \begin{aligned}
  \mathcal{E}_{\textit{ELLA}}  = \argminA_{\substack{\*L,\{\*s_1,\*s_2, \ldots \*s_T\}}}& \frac{1}{T}\sum_{t \in [T]} \mathcal{L}(\*y_t,f(\*X_t,\*L\*s_t)) \\
  & + \mu \Vert \*s_t \Vert_1  + \lambda \Vert \*L\Vert_F^2\\
  \end{aligned}
  \label{eq:ella}
  \end{equation}
  \end{comment}

   %%%%%%%%%%%%%%%%%%%%%%%%%%%%%%%%%%%%%%%%%%%%%%%%%%%%%%%%%%%%%%%%%%%%%%%%%

  \section{Experiments}

  %%%%%%%%%%%%%%%%%%%%%%%%%%%%%%%%%%%%%%%%%%%%%%%%%%%%%%%%%%%%%%%%%%%%%%%%%
All reported results in this section are averaged over $10$ random runs of the training data. Unless otherwise specified, all model parameters are chosen via $3$-fold cross validation. For all the experiments, we update the $\*\tau$ values using the equation \ref{eq:spmtlprob}. We evaluate our self-paced multitask learning algorithm on the four well-known multitask problems (MMTL, MTFL, MTASO), briefly discussed in the previous section. We also compare our results with Independent multitask learning (ITL) where each task is learned independently and Single-task learning (STL) where we learn a single model by pooling together data from all the tasks. 

\subsection{Synthetic Experiment}

\textbf{Synthetic data} \textit{(syn1)} consists of $30$ tasks that belong to $3$ groups of tasks with $15$ training examples per task. We generate the task parameters as in \cite{kang2011learning}. Each example consists of $20$ features. We randomly select a subset of tasks and increase their variance to $(\sigma=25)$, and variances for the rest of the tasks are set to be low $(\sigma=5)$ in order to simulate the difference between easy and hard tasks. With this setting, we expect that our self-paced learning algorithm should be able to learn the shared knowledge from the easier tasks and use this knowledge to improve the performance of the harder tasks.

\textbf{Synthetic data} \textit{(syn2)} consists of $30$ tasks with $15$ training examples per task as before. We randomly generate a $30$-dimensional vector $(s_1, s_2, s_3, \ldots, s_{30})$ such that the parameter for each task $t$ is given as $\*w_t=(s_1,s_2, \ldots s_t, 0, 0, \ldots, 0)$ and each example consists of $30$ features.  The dataset is constructed in such a way that learning the task $t$ is easier than learning the task $t+1$ and so on.

\begin{figure*}[ht]
	\centering
	\includegraphics[width=2.3in]{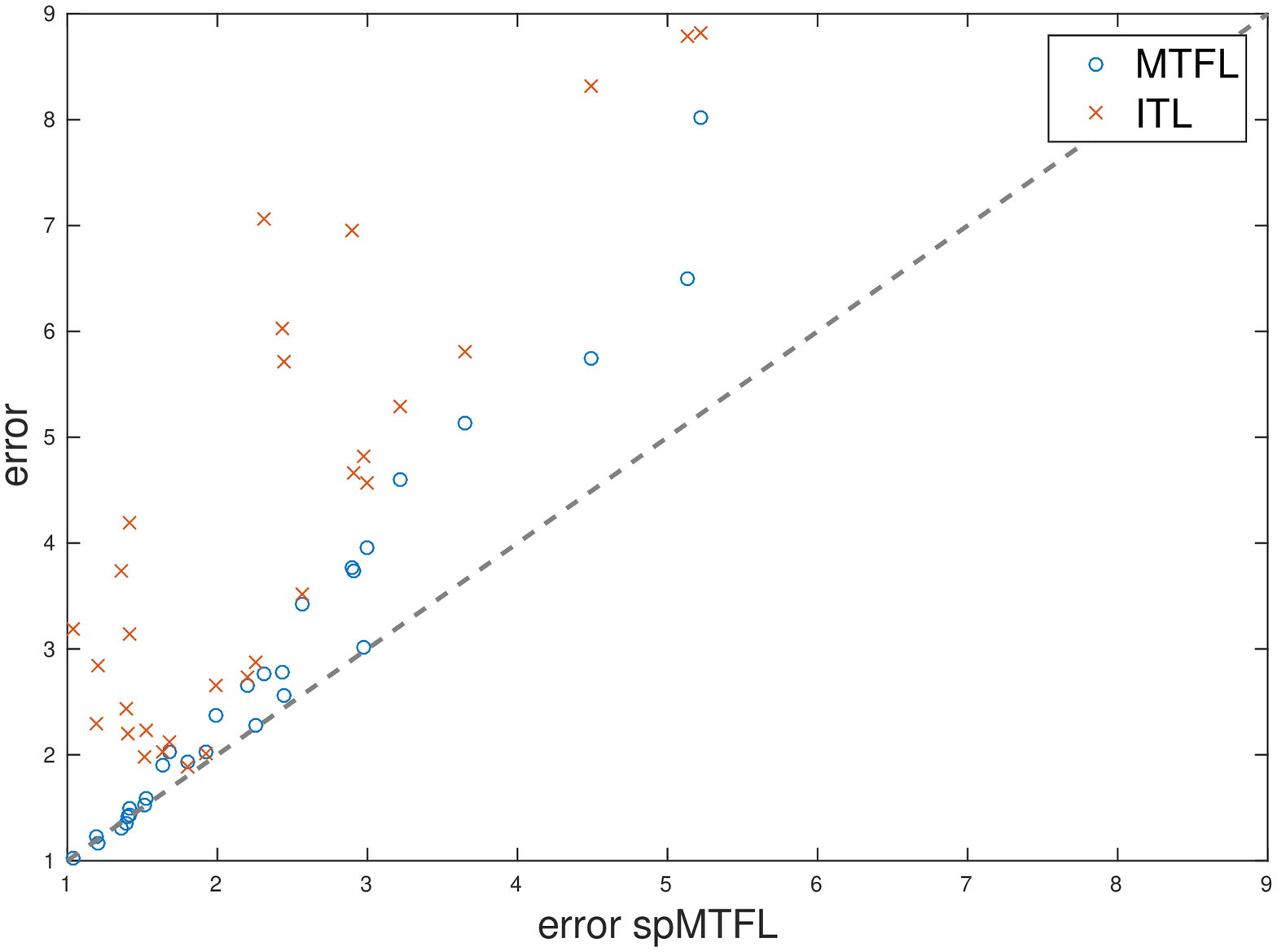}
	\includegraphics[width=2.3in]{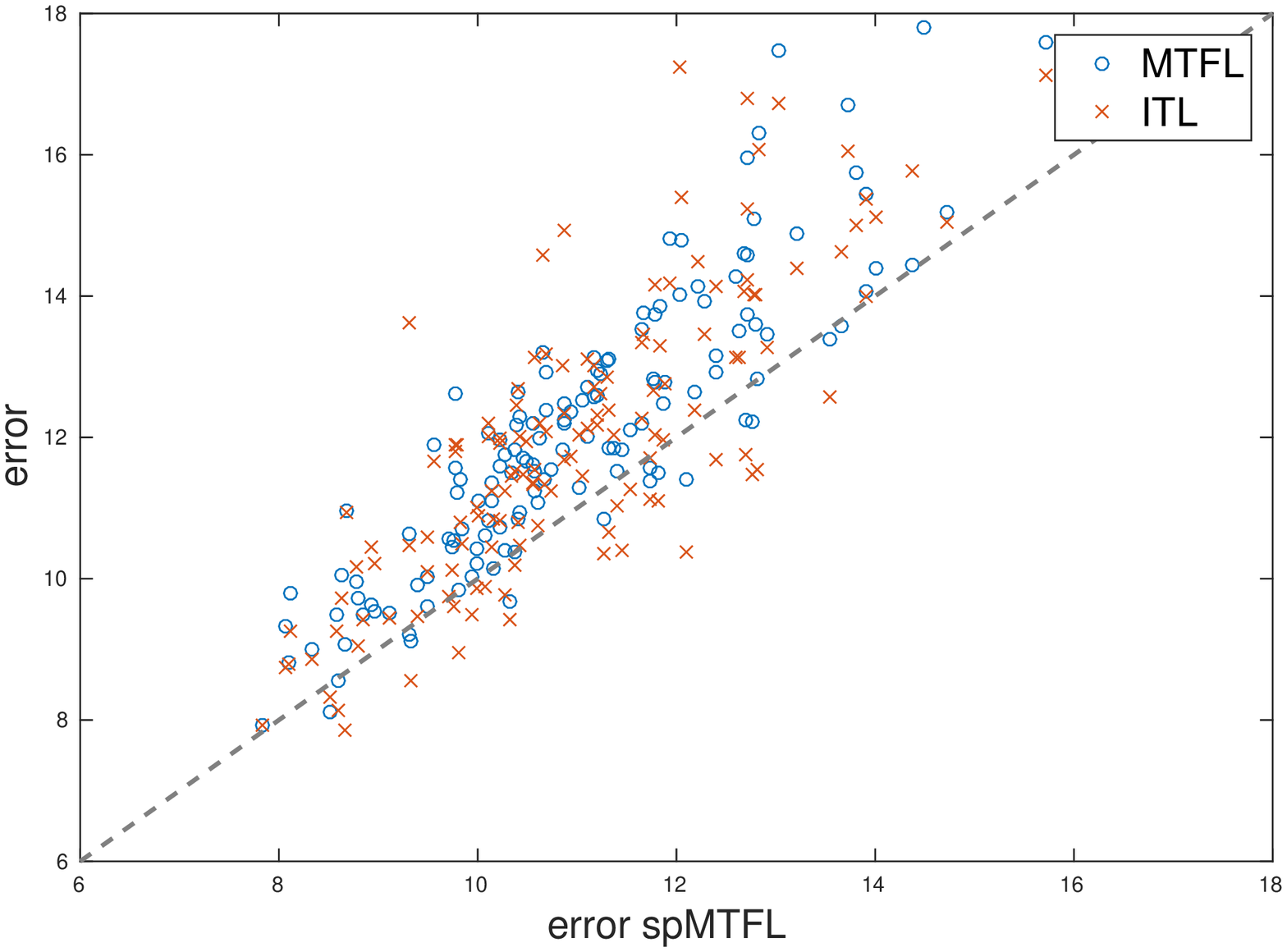}
	\includegraphics[width=2.3in]{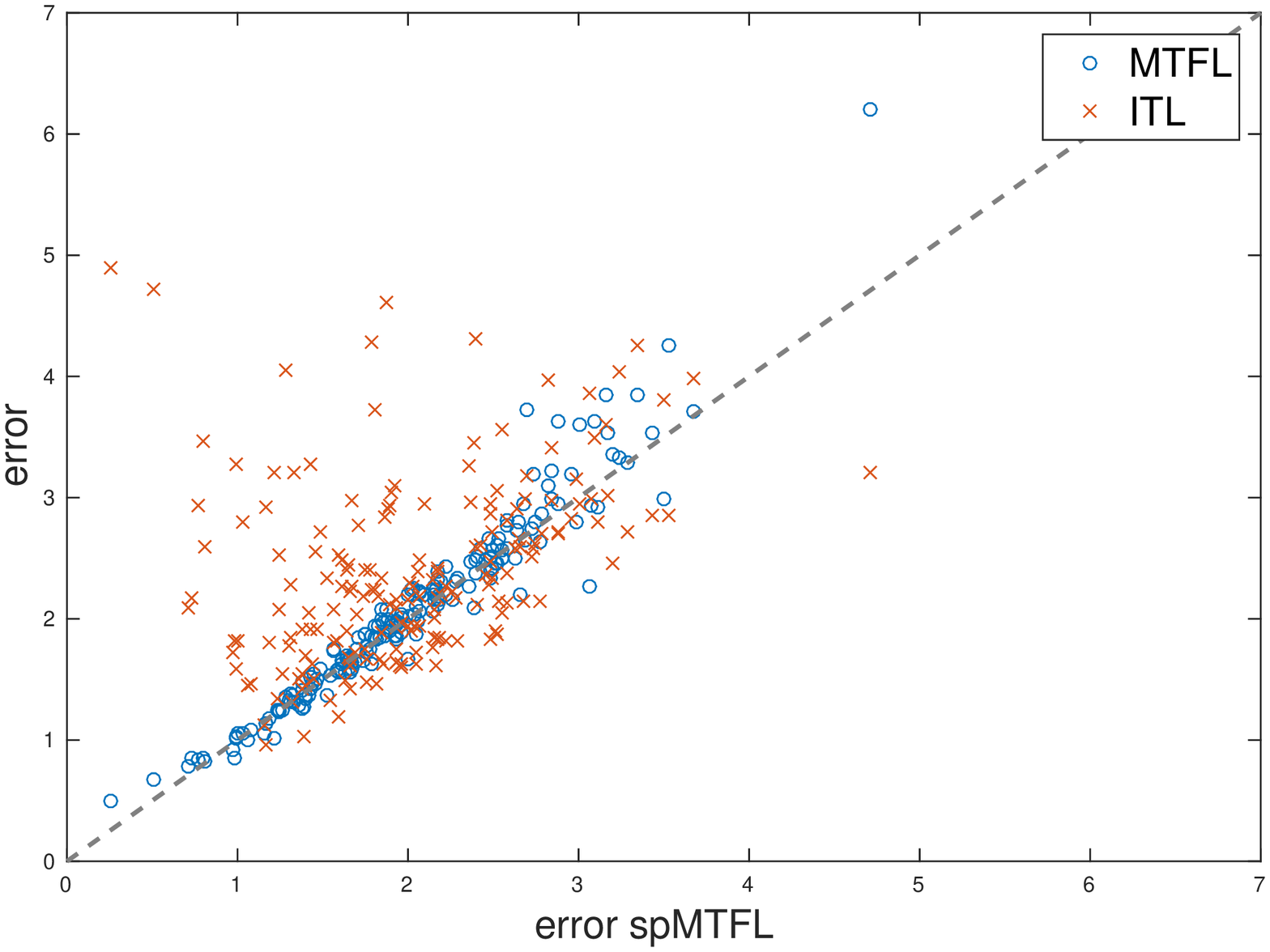}
	
	\includegraphics[width=2.3in]{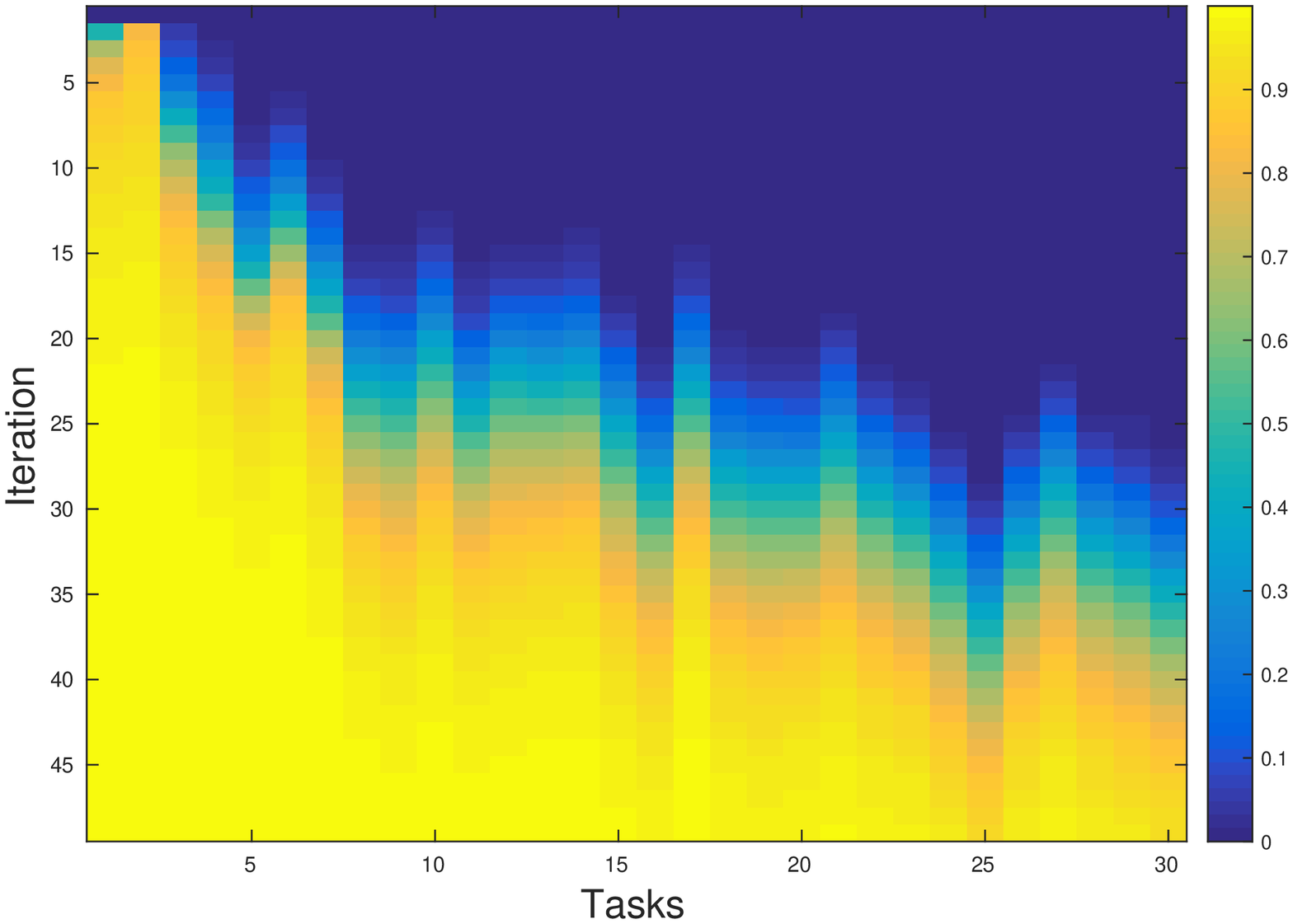}
	\includegraphics[width=2.3in]{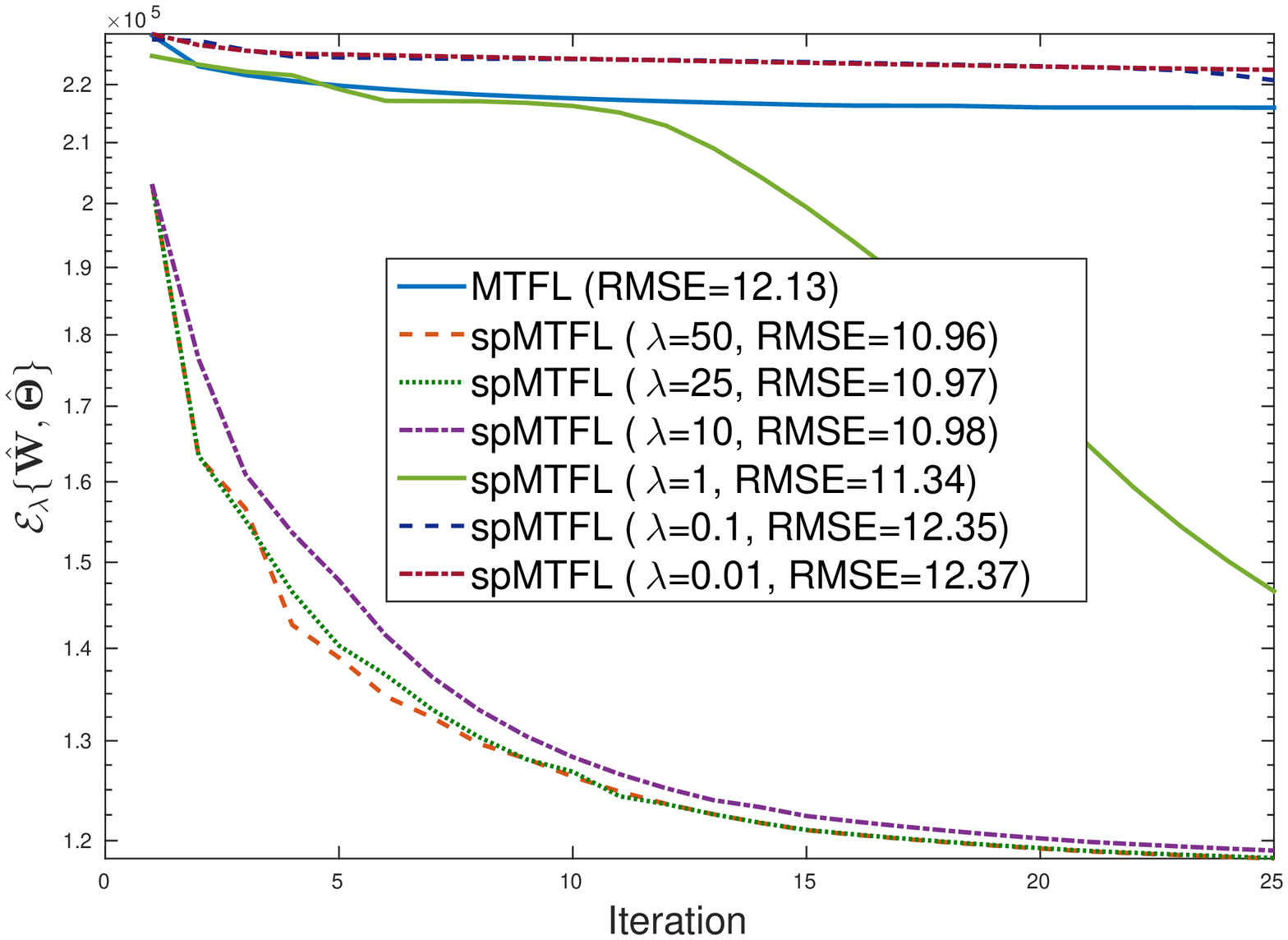}
	\includegraphics[width=2.3in]{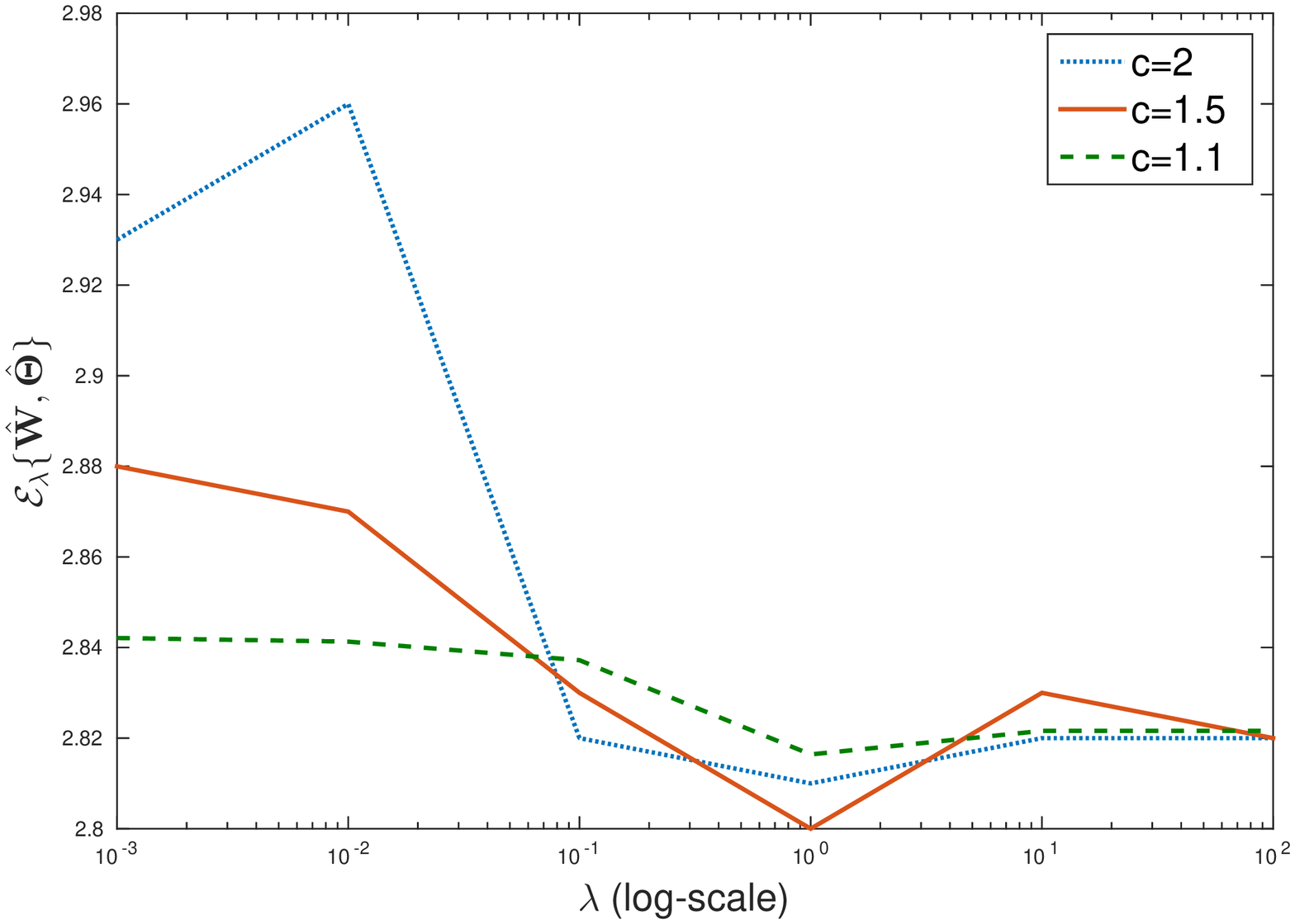}
	\caption{ Error of MTFL and ITL vs. Error of \textit{sp}MTFL calculated for \textit{syn2} dataset ({Top-left}).  Error of MTFL and ITL vs. Error of \textit{sp}MTFL calculated for \textit{school} dataset ({Top-middle}). Error of MTFL and ITL vs. Error of \textit{sp}MTFL calculated for \textit{cs} dataset ({Top-right}). Values of $\hat{\tau}$ from \textit{sp}MTFL at each iteration calculated for \textit{syn2} dataset ({Bottom-left}). Convergence of the algorithm with varying threshold $\lambda$ ({Bottom-middle}) calculated from \textit{sp}MTFL for \textit{school} dataset. Convergence of the algorithm with different learning pace $'c'$ ({Bottom-right}) calculated from \textit{sp}MTFL for \textit{cs} dataset. The experiment shows $'c'=1.1$ for learning pace yields a stable performance.}
	\label{fig:cloud}
\end{figure*}

\begin{table*}[ht]
\centering

	\renewcommand{\arraystretch}{1.5}
\begin{tabular}{|l|l|l|l|l|l|l|}
\hline
\textit{\textbf{Models}} & \textit{syn1}        & \textit{syn2}        & \textit{school}       & \textit{cs}          & \textit{sentiment}    & \textit{landmine}     \\ \hline
\textbf{STL}             & 1.60 (0.02)          & 4.16 (0.09)          & 12.13 (0.08)          & 2.45 (0.13)          & 58.49 (0.40)          & 74.11 (0.50)          \\ \hline
\textbf{ITL}             & 1.13 (0.07)          & 3.25 (0.10)          & 12.00 (0.04)          & 1.99 (0.14)          & 68.39 (0.34)          & 74.39 (1.11)          \\ \hline\hline
\textbf{MMTL}            & 1.12 (0.07)          & 3.24 (0.10)          & 12.10 (0.08)          & 1.99 (0.18)          & 68.54 (0.27)          & 75.50 (1.86)          \\ \hline
\textbf{spMMTL}          & \textbf{1.03 (0.05)} & 3.24 (0.10)          & \textbf{10.34 (0.06)} & 1.89 (0.10)          & 68.54 (0.26)          & 75.73 (1.29)          \\ \hline\hline
\textbf{MTFL}            & 0.81 (0.06)          & 2.82 (0.13)          & 12.06 (0.08)          & 1.91 (0.18)          & 68.91 (0.31)          & 75.67 (1.03)          \\ \hline
\textbf{spMTFL}          & \textbf{0.73 (0.05)} & \textbf{2.34 (0.12)} & \textbf{10.99 (0.08)} & 1.87 (0.15)          & \textbf{75.60 (0.17)} & \textbf{76.92 (1.06)} \\ \hline\hline
%\textbf{MTML}            & 0.79 (0.05)          & 2.66 (0.10)          & 11.27 (0.08)          & 1.16 (0.24)          & 69.73 (0.15)          & 75.97 (1.18)          \\ \hline
%\textbf{spMTML}          & 0.78 (0.04)          & 2.66 (0.13)          & 11.27 (0.08)          & \textbf{0.97 (0.18)} & 69.65 (0.14)          & 75.98 (1.04)          \\ \hline\hline
\textbf{MTASO}           & 0.56 (0.03)          & 2.66 (0.16)          & 11.14 (0.10)          & 1.38 (0.19)          & 72.03 (0.18)          & 72.58 (1.46)          \\ \hline
\textbf{spMTASO}         & 0.52 (0.03)          & \textbf{2.54 (0.14)} & 11.14 (0.11)          & \textbf{1.12 (0.17)} & 72.36 (0.19)          & \textbf{75.73 (1.46)} \\ \hline\hline
\end{tabular}
\caption{Average performance on six datasets: means and standard errors over $10$ random runs.  We use RMSE as our performance measure for \textit{syn1}, \textit{syn2}, \textit{school}, and \textit{cs} and Area under the curve (AUC) for \textit{sentiment} and \textit{landmine}. Self-paced methods with the best performance against their corresponding MTL baselines (paired t-tests at $95\%$ significance level) are shown in boldface.}
	\label{tab:results1}
\end{table*}

The result for \textit{syn1} and \textit{syn2} are shown in Table \ref{tab:results1}. We report the RMSE (mean and std) of our methods. All of our self-paced methods perform better than their baseline methods on average in both the synthetic datasets. Figure \ref{fig:cloud} (bottom-left) shows the $\tau$ learned using self-paced task selection (\textit{sp}MTFL) at each iteration. We can see that the tasks are selected based on their difficulty and the number of features used in each task.
Figure \ref{fig:cloud} (top-left) shows the task-specific test errors for \textit{syn2} dataset (\textit{sp}MTFL vs. their corresponding baseline methods MTFL and ITL). Each red point in the plot compares the RMSE of ITL  with  \textit{sp}MTFL and each blue point compares the RMSE of MTFL vs. \textit{sp}MTFL. Points above the line $y=x$ show that the self-paced methods does better than ITL or their MTL baseline methods. From the (MTFL vs. \textit{sp}MTFL) plot, we can see that our self-paced learning method \textit{sp}MTFL achieves significant improvement on harder tasks (blue points in top-right) compared to the easier tasks (blue points in bottom-left). Based on our learning procedure, these harder tasks must have been learned at the later part of the learning and thus efficiently utilize the knowledge learned from the easier tasks to improve their performances. Similar behaviour can be observed in the other two plots. Note that some of the points fall slightly below the $y=x$ line, but since the decrease in performance of these tasks are small, it has very little impact on the overall score. We believe this can be avoided if we tune different regularization parameter $\lambda_t$ for each task. However, this will increase the number of parameters to tune in addition to the task weight parameters $\tau$.

\subsection{Evaluation on Real Data}

\begin{comment}
\begin{table*}[h]
	\renewcommand{\arraystretch}{1.5}
	\centering
	\begin{tabular}{l|l|l|l|l|}
\cline{2-5}
\textbf{}                     & \textit{syn}   & \textit{school} & \textit{cs}    & \textit{sarcos} \\ \hline
\multicolumn{1}{|l|}{STL}     & 1.096 $\pm$ 0.060 & 11.72 $\pm$ 0.15   & 2.390 $\pm$ 0.222 & 0.2563 $\pm$ 0.004 \\ \hline
\multicolumn{1}{|l|}{MTFL}    & 0.639 $\pm$ 0.064 & 10.18 $\pm$ 0.12   & 1.954 $\pm$ 0.097 & 0.1321 $\pm$ 0.002 \\
\multicolumn{1}{|l|}{\textit{sp}MTFL}  & \textbf{0.579} $\pm$ 0.042 & 10.17 $\pm$ 0.13   & \textbf{1.758} $\pm$ 0.127 & \textbf{0.1247} $\pm$ 0.002 \\ \hline
\multicolumn{1}{|l|}{MTML}    & 0.753 $\pm$ 0.044 & 10.34 $\pm$ 0.16   & 1.858 $\pm$ 0.086 & 0.1349 $\pm$ 0.002 \\
\multicolumn{1}{|l|}{\textit{sp}MTML}  & \textbf{0.721} $\pm$ 0.037 & 10.34 $\pm$ 0.15   & 1.838 $\pm$ 0.093 & 0.1346 $\pm$ 0.002 \\ \hline
\multicolumn{1}{|l|}{MTASO}   & 0.778 $\pm$ 0.063 & 10.2 $\pm$ 0.12   & 2.237 $\pm$ 0.286 & 0.1260 $\pm$ 0.001 \\
\multicolumn{1}{|l|}{\textit{sp}MTASO} & \textbf{0.748} $\pm$ 0.015 & 10.2 $\pm$ 0.13   & \textbf{1.808} $\pm$ 0.091 & 0.1256 $\pm$ 0.002 \\ \hline
\end{tabular}
	\caption{Average performance on three datasets: means and standard errors over $10$ random runs.  We use RMSE as our performance measure for \textit{school}, \textit{cs} and \textit{sarcos}. Self-paced methods with the best performance (paired t-tests at $95\%$ significance level) are shown in boldface}
	\label{tab:results1}
\end{table*}
\end{comment}
We use the following benchmark real datasets for our experiments on self-paced multitask learning. 

 \textbf{London School data} (\textit{school}) consists of examination scores of $15,362$ students from $139$ schools in London. Each school is considered as a task and the feature set includes  year of the examination, four school-specific and three student-specific features. We replace each categorical feature with one binary variable for each possible feature value, as suggested in \cite{argyriou2008convex}. This results in $26$ features with additional feature to account for the bias term. We use the ten $20\%-80\%$ train-test splits that came with the dataset for our experiments.

\textbf{Computer Survey data} (\textit{cs}) was collected from the ratings of $190$ students on each of the $20$ different personal computers. Each student here is considered as a single task and the rating ranges from $0-10$. There are $20$ observations in each task. Each computer is represented by $13$ different features such as RAM, cache-size, CPU speed, etc. We add an additional feature to account for the bias term.  Train-test splits are obtained by selecting $75\%-25\%$, thus giving $15$ examples for training and $5$ examples for test set.

\textbf{Sentiment Detection data} (\textit{sentiment}) contains reviews from $14$ domains. The reviews are represented by a bag of unigram/bigram \textit{TF-IDF} features from a dictionary of size $28,775$. Each review is associated with a rating from $\{1,2,4,5\}$. We select $1,000$ reviews for each domain and create two tasks ($500$ reviews per task), based on whether the rating is $5$ or not and whether the rating is $1$ or not, in order to represent the different levels of  sentiment. This gives us $28$ binary classification tasks. We use $120$ reviews per task for training and the rest of the reviews for test set.

 \textbf{Landmine Detection data} (\textit{landmine}) consists of $19$ tasks collected from different landmine fields. Each task is a binary classification problem: landmines $(+)$ or clutter $(-)$ and each example consists of 9 features extracted from radar images. Landmine data is collected from two different terrains:  tasks 1-10 are from highly foliated regions and tasks 11-19 are from desert regions, therefore tasks naturally form two clusters.  We use $80$ examples from each task for training and the rest as the test data. We repeat the experiments on $10$ (stratified) splits to measure the performance reliably. Since the dataset is highly skewed, we use $AUC$ score to compare our results.

\begin{comment}
\begin{figure*}[ht]
	\centering
	\includegraphics[width=2in]{img/syn1-mtfl}
	\hspace{0.5em}
	\includegraphics[width=2in]{img/syn1-mtfl}
    \hspace{0.5em}
	\includegraphics[width=2in]{img/syn1-mtfl}
	
	\caption{\textbf{Top}: Sensitive analysis for the regularization parameter $\lambda_2$ when $\lambda_1=0.1$ and $K=2$ (left) and number of clusters $K_1$ and $K_2$ when $\lambda_1=0.1$ and $\lambda_2=0.1$ (right) calculated for \textit{syn5} dataset.  \textbf{Middle}: Mean accuracy (left) and runtime (right) calculated for \textit{Oxford} dataset with varying training set sizes.}
	\label{fig:obj}
\end{figure*}
\end{comment}
Table \ref{tab:results1} summarizes the performance of our methods on the four real datasets. We can see that our proposed self-paced learning algorithm does well on almost all datasets. As in our synthetic experiments, we observe that \textit{sp}MTFL performs significantly better than MTFL, which is a state-of-the-art method for multitask problems. It is interesting to see that when the self-paced learning procedure doesn't help the original algorithm, it doesn't perform worse than the baseline results. In such cases, our self-paced learning algorithm gives equal probability to all the tasks $(\tau_t= \frac{1}{T}, \forall t \in [T])$ within the first few iterations. Thus the proposed self-paced methods reduce to their original methods and the performance of the self-paced methods are on par with their baselines. 

We also notice that if a dataset doesn't adhere to the assumptions of a model, such as task parameters lie on a manifold or low-dimensional space, then our self-paced methods result in little improvement, as it can be seen in \textit{cs} (and also in \textit{sentiment} for \textit{sp}MTASO). %Since \textit{school} dataset does not have a low-dimensional representation, our self-paced algorithm performs equally well compared to the original methods. 
It is worth mentioning that our proposed self-paced multitask learning algorithm does exceptionally better in \textit{school}, which is a benchmark dataset for multitask experiments in the existing literature \cite{agarwal2010learning,kumar2012learning}. Our proposed methods achieve as much as $14\%$  improvement over their baselines on some experiments.
 Figures (top-middle) and (top-right) show the task-specific errors for \textit{school} and \textit{cs} dataset. We can see similar pattern as in \textit{syn2}. The easier tasks learned at an earlier stage help the harder tasks at the later stages as it is evident from these plots.

 %Figure (middle-bottom) shows the convergence of the algorithm for the different values of $\lambda$. We can see that for a specific value of $\lambda$, the self-paced learning procedure (\textit{sp}MTFL) converges much faster than the baseline \textit{MTFL}. Figure (right-bottom) shows the comparison of error for different values of learning pace parameter $'c'$. It is evident from the plot that we get a stable and consistent results when we choose $c=1.1$, which is the value used in all of our experiments.

 \subsection{Comparing \textit{sp}MTFL with Sequential Learning Algorithms}
 
 Finally, we compare our self-paced multitask learning algorithm against the sequential multitask learning algorithms (curriculum learning for multiple tasks \cite{pentina2015curriculum} and efficient lifelong learning \cite{ruvolo2013ella,ruvolo2013active} \footnote{\url{http://www.seas.upenn.edu/~eeaton/software/ELLAv1.0.zip}}. We choose \textit{sp}MTFL for comparison based on its overall performance in the previous experiments. We use landmine dataset for evaluation. We use different variant of \textit{ELLA} for fair comparison against our proposed approach. The original \textit{ELLA} algorithm assumes that the tasks arrive randomly and the lifelong learner has no control over their order (\textit{ELLA-random}). \citeauthor{ruvolo2013active} (\citeyear{ruvolo2013active}) show that if the learner can choose the next task actively, it can improve the learning performance using as few tasks as possible. They proposed two active task selection procedures for choosing the next best task: 1) Information Maximization (\textit{ELLA-infomax}) chooses the next task to maximize the expected information gain about the basis $\*L$; 2) Diversity (\textit{ELLA-diversity}) chooses the next task as the one that the current basis $\*L$ is doing the worst performance. Both these approaches select the tasks that are significantly different from the previously learned tasks (\textit{active task selection}), rather than a progression of tasks that build upon each other. Our proposed method selects the task based on the training error and its relevance  to the shared knowledge learned from the previous tasks (\textit{self-paced task selection}).

\begin{figure}[ht]
	\centering
	\includegraphics[width=3in]{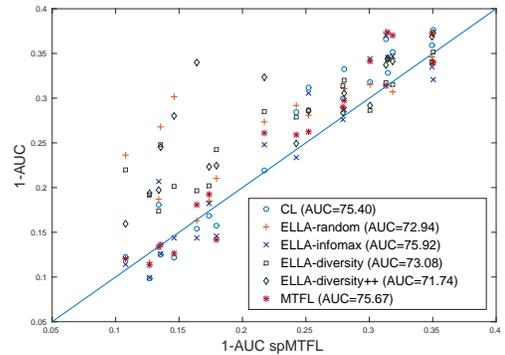}
	\caption{ Average performance on \textit{landmine} for sequential learning algorithms and \textit{sp}MTFL: means and standard errors over $10$ random runs.  We use $(1-AUC)$ score as our performance measure for comparison. Mean $AUC$ score is shown in the bracket.}
	\label{fig:results2}
\end{figure}
 
\begin{comment}
\begin{table}[ht]
	\renewcommand{\arraystretch}{1.5}
	\centering
	\begin{tabular}{l|l|}
\cline{2-2}
  \textbf{}          & \textit{landmine} \\ \hline
\multicolumn{1}{|l|}{ITL}               & 74.89 $\pm$ 0.50      \\ \hline
\multicolumn{1}{|l|}{CL}                & 75.40 $\pm$ 1.41      \\ \hline\hline
\multicolumn{1}{|l|}{ELLA-random}       & 72.94 $\pm$ 1.87      \\ \hline
\multicolumn{1}{|l|}{ELLA-infomax}      & 75.92 $\pm$ 1.87      \\ \hline
\multicolumn{1}{|l|}{ELLA-diversity}    & 73.08 $\pm$ 1.84      \\ \hline
\multicolumn{1}{|l|}{ELLA-diversity++} & 71.74 $\pm$ 1.80      \\ \hline\hline
\multicolumn{1}{|l|}{MTFL}              & 75.67 $\pm$ 1.03      \\ \hline
\multicolumn{1}{|l|}{\textit{sp}MTFL}            & \textbf{76.92 $\pm$ 1.05}      \\ \hline
\end{tabular}
	\caption{Average performance on landmine: means and standard errors over $10$ random runs.  We use $AUC$ score as our performance measure for \textit{landmine}.}
	\label{tab:results2}
\end{table}
\end{comment}

Figure \ref{fig:results2} shows the task-specific test performance results for this experiment on \textit{landmine} dataset. We compare our results from \textit{sp}MTFL against \textit{CL} and variants of \textit{ELLA}. We use $(1-AUC)$ score for our comparison. As in Figure \ref{fig:cloud}, points above the line $y=x$ show that the \textit{sp}MTFL does better than the other sequential learning methods. We can see that \textit{sp}MTFL outperforms all the baselines on average ($76.92$). Compared to \textit{sp}MTFL, \textit{CL} performs better on easier tasks but worse on harder tasks. On the other hand, the performance of the variants of \textit{ELLA} on harder tasks are comparable to that of our self-paced method, but worse on some easier tasks.

%It is interesting to see that the lifelong learning algorithm with active task selection does worse than even ITL for some task selection strategy. It is because ELLA learns the tasks in an online fashion and does not update the previously learned tasks whenever $\*L$ is updated. This is less of an issue if the tasks have considerable amount of data, but since we use limited data for our experiment, it hurts the performance of the lifelong learning algorithm significantly.
  %%%%%%%%%%%%%%%%%%%%%%%%%%%%%%%%%%%%%%%%%%%%%%%%%%%%%%%%%%%%%%%%%%%%%%%%%

  \section{Conclusion and Future Work}

  %%%%%%%%%%%%%%%%%%%%%%%%%%%%%%%%%%%%%%%%%%%%%%%%%%%%%%%%%%%%%%%%%%%%%%%%%

In this work, we proposed a novel self-paced learning framework for multiple tasks that jointly learns the latent task weights and shared knowledge from all the tasks. The proposed method iteratively updates the shared knowledge based on these task weights and thus improves the learning performance. By allowing the $\*\tau$ to take the probabilistic interpretation, we can easily see which tasks are easier to learn at any iteration, and prefer those for task selection. In our future work, we plan to consider a stochastic version of this algorithm to update the shared knowledge base efficiently and study the algorithm's ability to handle the outlier tasks. Effectiveness of our algorithm is empirically verified over several benchmark datasets.
\newpage
 \bibliography{references}
 \bibliographystyle{named}
  \end{document}